\documentclass[letterpaper]{article} 
\usepackage{aaai2026}  
\usepackage{times}  
\usepackage{helvet}  
\usepackage{courier}  
\usepackage[hyphens]{url}  
\usepackage{graphicx} 
\usepackage{natbib}  
\usepackage{caption} 
\usepackage{algorithm}
\usepackage{algorithmic}

\usepackage{multirow}
\usepackage{amsmath}
\usepackage{comment}
\usepackage{booktabs} 
\usepackage{colortbl} 
\usepackage{xcolor}   
\usepackage{enumerate}
\usepackage{newfloat}
\usepackage{listings}
\usepackage{float}
\usepackage{appendix}

\DeclareCaptionStyle{ruled}{labelfont=normalfont,labelsep=colon,strut=off} 
\floatstyle{ruled}
\newfloat{listing}{tb}{lst}{}
\floatname{listing}{Listing}
\title{PixCLIP: Achieving Fine-grained Visual Language Understanding via Any-granularity Pixel-Text Alignment Learning}
\author{
    Yicheng Xiao\textsuperscript{\rm 1,2}\equalcontrib, Yu Chen\textsuperscript{\rm 1,2}\equalcontrib, Haoxuan Ma\textsuperscript{\rm 3}\equalcontrib, Jiale Hong\textsuperscript{\rm 4}, Caorui Li\textsuperscript{\rm 5}, Lingxiang Wu\textsuperscript{\rm 1,2}, Haiyun Guo\textsuperscript{\rm 1,2}, Jinqiao Wang\textsuperscript{\rm 1,2}\thanks{Corresponding Author}
}
\affiliations{
    \textsuperscript{\rm 1}CASIA,
    \textsuperscript{\rm 2}UCAS,
    \textsuperscript{\rm 3}NJU,
    \textsuperscript{\rm 4}SJTU,
    \textsuperscript{\rm 5}SEU

}

\makeatletter
\providecommand{\copyright@text}{}
\newif\ifcopyright@on
\copyright@onfalse
\makeatother

\begin{document}

\maketitle

\begin{abstract}
While the Contrastive Language-Image Pretraining(CLIP) model has achieved remarkable success in a variety of downstream vison language understanding tasks, enhancing its capability for fine-grained image-text alignment remains an active research focus.  To this end, most existing works adopt the strategy of explicitly increasing the granularity of visual information processing, e.g., incorporating visual prompts to guide the model focus on specific local regions within the image. Meanwhile, researches on Multimodal Large Language Models(MLLMs) have demonstrated that training with long and detailed textual descriptions can effectively improve the model's fine-grained vision-language alignment. However, the inherent token length limitation of CLIP's text encoder fundamentally limits CLIP to process more granular textual information embedded in long text sequences. To synergistically leverage the advantages of enhancing both visual and textual content processing granularity, we propose PixCLIP, a novel framework designed to concurrently accommodate visual prompt inputs and process lengthy textual descriptions. Specifically, we first establish an automated annotation pipeline capable of  generating pixel-level localized, long-form textual descriptions for images. Utilizing this pipeline, we construct LongGRIT, a high-quality dataset comprising nearly 1.5 million samples. Secondly, we replace CLIP's original text encoder with the LLM and propose a three-branch pixel-text alignment learning framework, facilitating fine-grained alignment between image regions and corresponding textual descriptions at arbitrary granularity. Experiments demonstrate that PixCLIP showcases breakthroughs in pixel-level interaction and handling long-form texts, achieving state-of-the-art performance on fine-grained pixel-level tasks, while retaining the capabilities of original CLIP in handling global-level vision-language tasks.

\end{abstract}
\begin{links}
    \link{Code}{https://github.com/StuHude/PixCLIP/}
\end{links}

\section{Introduction}

\begin{figure*}
  \centering
  \includegraphics[width=\textwidth]{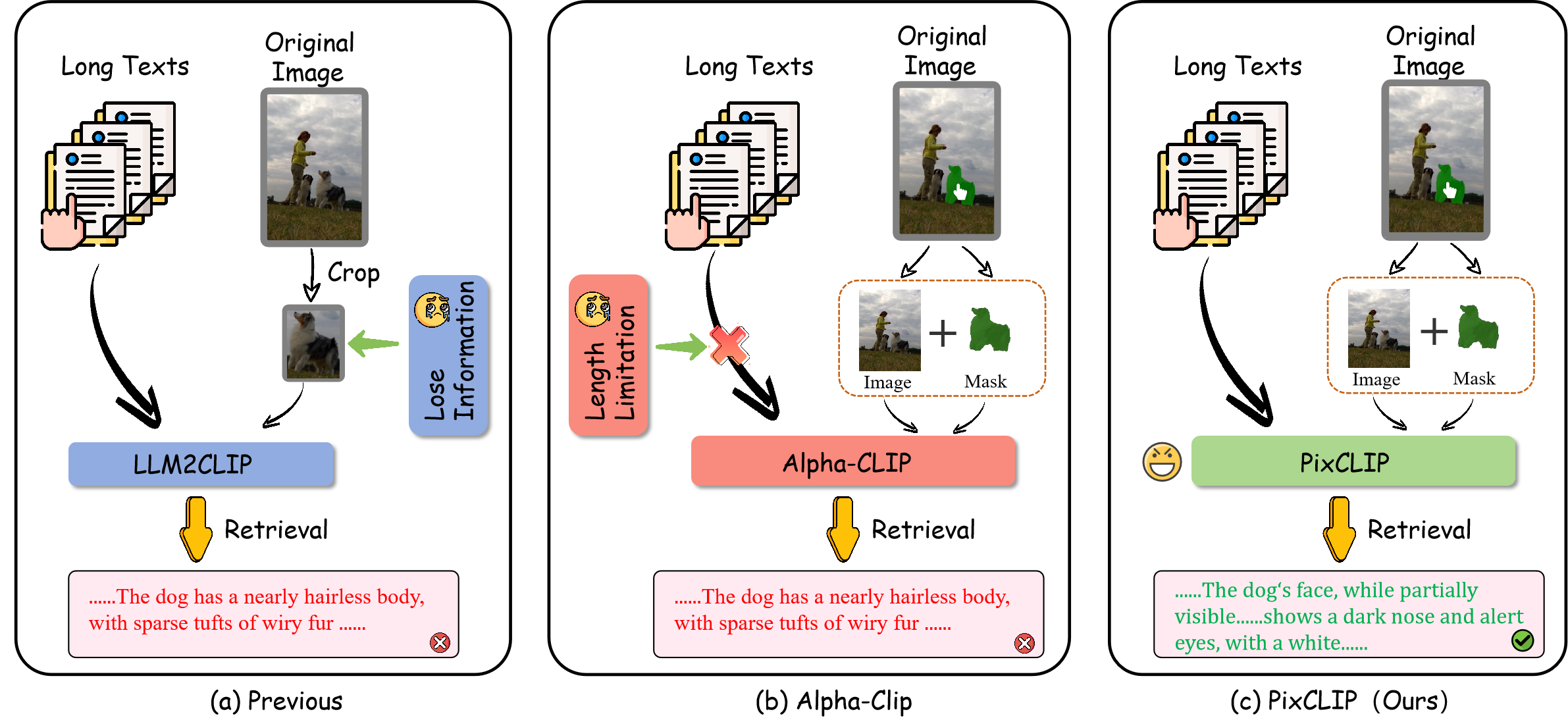}
  \caption{As shown on the (a), prior methods fail on fine-grained local tasks due to inability to accept mask input, While existing promptable methods struggle to handle long texts due to token limitations, as shown in (b). Only our method succeeds in yielding robust embeddings for image and text at any granularity.}
  \label{fig:first}
\vspace{-12pt}
\end{figure*}

Contrastive Language-Image Pretraining (CLIP) \cite{radford2021learningtransferablevisualmodels} is a multimodal model that jointly trains image and text encoders via contrastive learning to align their features in a shared semantic space. Its features aim to capture all the semantic details within images,  demonstrating strong representation capabilities and exceptional generalizability. This inherent versatility makes them suitable for a variety of downstream tasks, such as open-world recognition \cite{chen2023ovarnet, dong2023maskclip, lan2024clearclip}, retrieval \cite{huang2024llm2clip, saito2023pic2word, zhang2024long} and Multimodal Large Language Models (MLLMs) \cite{liu2023visualinstructiontuning,yang2024lever}.

Although CLIP captures the content of the entire image, focusing on regions of interest is crucial for achieving a more refined understanding. Traditional image processing techniques, such as cropping regions or masking irrelevant parts of images, tend to destroy and overlook contextual information. As a result, recent works have focused on enhancing the regional representation of CLIP, enabling it to accept visual prompts that guide attention to specific regions of interest. Visionllm v2 \cite{wu2024visionllm}, ControlMLLM \cite{wu2024controlmllm}, and Ferret \cite{you2023ferretrefergroundgranularity} have introduced prompt encoders to enable multimodal models to receive visual prompt inputs, but these models cannot directly obtain visual representations to perform tasks such as retrieval. CLOC \cite{chen2025contrastivelocalizedlanguageimagepretraining} achieves alignment through a large number of bounding box-text pairs. However, its capability is limited to bounding boxes, not masks, and thus fails to capture the fine-grained, pixel-level details that masks can offer. Alpha CLIP \cite{sun2024alpha} has addressed these issues by adding an alpha channel and using mask-text pairs for contrastive learning. However, since the training focused primarily on short and concise captions, these methods tend to struggle with longer and more detailed textual descriptions, performing poorly on tasks that require fine-grained analysis. So, is it possible to build a model that overcomes the limits to accept visual prompts for local regions and complex long texts, achieving image-text alignment at arbitrary granularity?

In this paper, we propose PixCLIP, a universal CLIP model for fine-grained understanding through any-granularity pixel–text alignment learning, as shown in Fig.\ref{fig:first}. We innovatively perform pixel-level semantic alignment between masks and complex long texts. After fine-tuning, the model can dynamically embed any local region of an image and align it with complex texts of arbitrary length. PixCLIP achieves a breakthrough in both the visual and textual domains, surpassing traditional granularity limits, and establishes itself as a more fine-grained foundational model.

Specifically, existing mask-text datasets like GRiT 
 \cite{peng2023kosmos} and FERRET 
 \cite{you2023ferretrefergroundgranularity} are limited by short text and a lack of large-scale, well-aligned training data. To address this, we first propose \textbf{LongGRIT}, an automatically generated dataset of 1.5 million image-local fine-grained expressions designed to boost model performance. We also utilized 2,000 masked regions from LongGRIT to benchmark existing models for fine-grained local mask-text retrieval in complex settings, uncovering significant opportunities for improvement and further research.

Secondly, we propose a multi-grained alignment framework.    (1)We use an LLM as the text encoder and then employ contrastive learning with mask-text pairs.    This enables the model to directly obtain embeddings from images with mask and align them with fine-grained text features.    (2)In parallel, we utilize Fine-grained Cropping Alignment.    This branch extracts features solely from the region indicated by the mask.    It aims to make these features more similar to the features from the first step , thereby enhancing the features to concentrate more strongly on the masked region.    (3)Simultaneously, in the Local-Global Representation Enhancement branch, the mask covers the whole image, equivalent to extracting whole-image features and then mutually enhances the initial mask embedding by features from the corresponding region, jointly optimizing the representation capabilities for both the global and local.

We conduct extensive experiments on RefCOCO \cite{lin2015microsoftcococommonobjects}, Ref-SAV \cite{yuan2025sa2va}, Instance-COCO \cite{lin2015microsoftcococommonobjects}, ImageNet-S \cite{Gao_2023} for region-level evaluation, as well as a series of image-text retrieval datasets such as DOCCI, Urban, and Flickr. The results demonstrate that our model successfully enables the processing of complex long texts while maintaining the ability to accept local visual cues, greatly improving the performance of fine-grained tasks.

Our contribution is threefold:
\begin{enumerate}
  \setlength{\itemsep}{0pt}
  \setlength{\parsep}{0pt}
  \setlength{\parskip}{0pt}
  \item We propose PixCLIP, a universal foundation model that extends the input to any local region and long texts. Extensive experiments demonstrate the superiority of this model in both region-level tasks and traditional tasks.
  \item We construct and release the LongGRIT dataset, which contains 1.5 million detailed mask descriptions automatically generated and validated by SOTA MLLMs.
  \item We propose a three-branch pixel-text alignment learning framework that significantly enhances the model's performance through fine-grained cropping and local-global enhancement.
\end{enumerate}

\section{Related Works}

\subsection{Localized Image Representation} To enable CLIP to disentangle regions from the whole image for more targeted processing and understanding, various methods have been explored in the field of segmentation. MaskCLIP \cite{dong2023maskclip} and ODISE \cite{xu2023openvocabularypanopticsegmentationtexttoimage} use attention masks to make CLIP focus more on local regions. Another approach is to change the input image by simply cropping or masking the image to leave only the foreground object. ReCLIP \cite{subramanian2022reclip} and OvarNet \cite{chen2023ovarnet} crop the original image using bounding box from object proposal network. However, the valuable context information is lost except for using complex post-process proposed in ReCLIP \cite{subramanian2022reclip}. Some other approaches prompt the CLIP by modifying the input image, guiding CLIP to focus on the area of interest. For example, Red-Circle \cite{shtedritski2023does}, FGVP \cite{yang2023finegrainedvisualprompting} use a circle or mask contour to tell CLIP where to focus. But the direct modification of images causes a domain gap with CLIP pertaining images. Alpha-CLIP \cite{sun2024alpha} incorporates an additional alpha channel,which proved that it is a more potential paradigm.

\subsection{Enabling Long Text for CLIP} It has been widely recognized that the quality of CLIP's text embedding is coarse and limited to only 77 tokens. Many works have attempted to extend the length of CLIP captions and retrain CLIP accordingly. DreamLIP \cite{zheng2024dreamliplanguageimagepretraininglong} leveraged ShareCaptioner \cite{chen2024sharegpt4videoimprovingvideounderstanding} and InstructBLIP to augment 30M captions. LongCLIP processes long text by encoding chunks and aggregating them again. These methods were compromised by splitting captions into multiple shorter segments \cite{zheng2024dreamliplanguageimagepretraininglong,fan2023improving}, or fine-tuning the positional encoding to support longer token inputs \cite{zhang2024long}. Recent work LLM2CLIP \cite{huang2024llm2clip} also explores aligning fine-tuned LLM with vision encoder directly. Our model uses LLM as the text encoder and successfully supports the ability for any-granularity visual representation while supporting long text inputs.

\subsection{Region-level image annotation} CLIP is pretrained on large-scale datasets like LAION-400M \cite{schuhmann2021laion400mopendatasetclipfiltered} and LAION-5B \cite{schuhmann2022laion5bopenlargescaledataset}, while fine-grained pixel-level labels are not available due to high manual labor costs. CLOC \cite{chen2025contrastivelocalizedlanguageimagepretraining} also generates fine-grained text labels via the pseudo-labeling pipeline. However, its data format is limited to box and cannot achieve pixel-level annotation, namely mask.Alpha-CLIP uses SAM \cite{kirillov2023segment} and BLIP \cite{li2022blip} to generate 20M object-level caption.But their local captions are either segmented from the entire image caption or generated by the clip-based captioning model, resulting in:(1) It is difficult to ensure the quality of data; (2) All region caption are only composed of several words, with limited length and insufficient fine-grained information. We proposed a fine-grained region-level image annotation framework, which obtained a large number of high-quality fine-grained long texts and region pairs from \textbf{multiple perspectives}, by using \textbf{multiple state-of-the-art MLLMs} and undergoing \textbf{multiple verifications}.

\section{Method}
\label{gen_inst}

\subsection{Fine-grained Data Generation}\label{sec:data_generation}
As illustrated in Fig.\ref{fig:data_pipeline}, our pipeline builds upon the GRIT-20M \cite{peng2023kosmos} dataset and consists of three stages: Object-level expression annotation, Context-level expression annotation, and Fine-grained expression annotation. Each stage progressively refines the target object's description, from appearance attributes to spatial context, culminating in a detailed referring expression.

\noindent\textbf{Object-level expression annotation.}
Initially, we isolate each object using its segmentation mask from GRIT-20M. Both the cropped object and its mask are fed into InternVL2-76B \cite{chen2025expandingperformanceboundariesopensource} to generate an initial {\em object-level} caption focused on visual attributes (e.g., shape, color, texture). To ensure semantic consistency, we employ Qwen2-72B \cite{yang2024qwen2technicalreport} to validate the caption against the original image. Any captions flagged as inconsistent (e.g., describing non-existent features) or semantically invalid (e.g., "a red clock" for a blue vase) are discarded, while valid captions are retained as {\em object-level} annotations.

\noindent\textbf{Context-level expression annotation.}
In this stage, we input the original image and the object's bounding box into Griffon-G-26B \cite{zhan2024griffongbridgingvisionlanguagevisioncentric}, a vision-language model specialized in spatial reasoning. The model generates a {\em context-level} caption that describes the object’s location (e.g., "leftmost on the mantelpiece") and its relationship to surrounding objects. This complements the object-level caption by capturing contextual information in the first stage.

\begin{figure*}[htb]
  \centering
  \includegraphics[width=\textwidth]{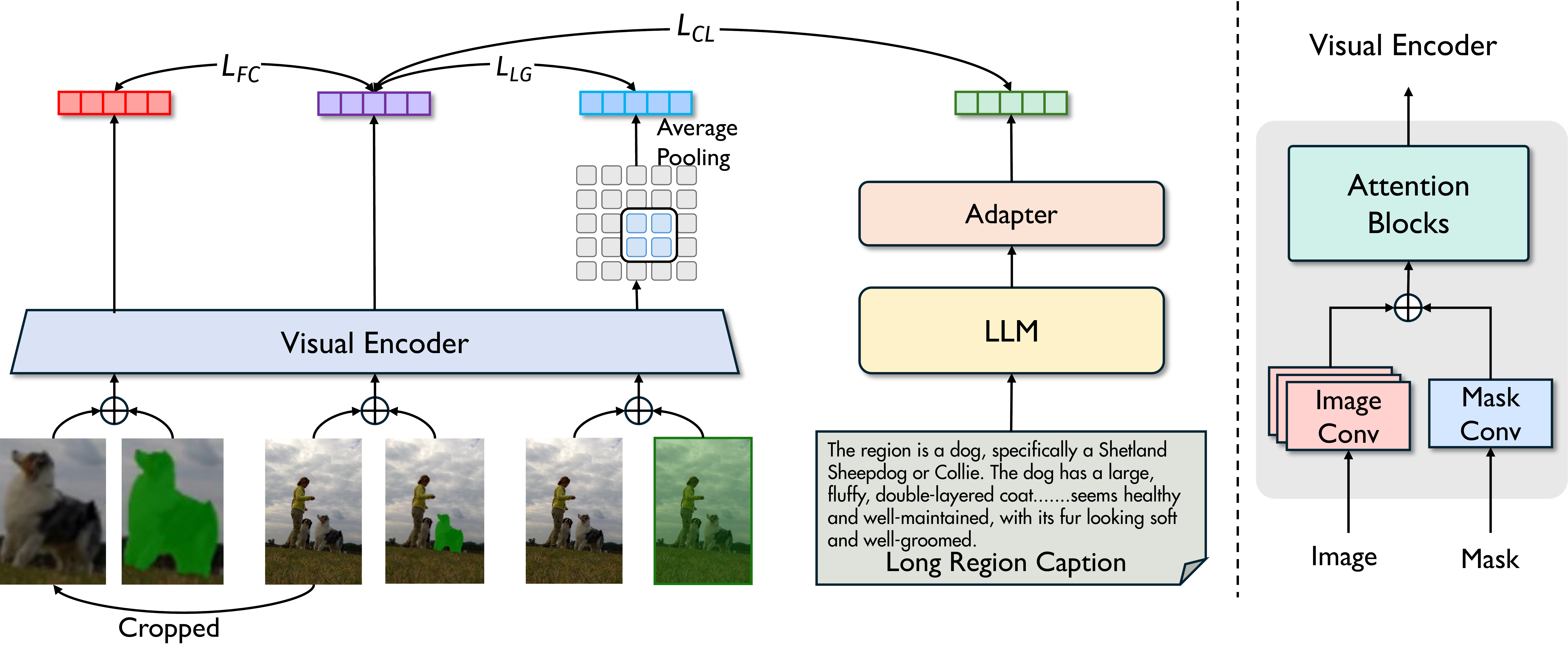}
  \caption{The framework of the proposed PixCLIP. Our model structure allows for inheriting weights from existing model and we propose the Fine-grained Cropping Alignment and Local-Global Representation Enhancement branches to enhance the mask-based visual embeddings.}
  \label{pipeline}
\end{figure*}

\noindent\textbf{Fine-grained expression annotation.}
Finally, we integrate the object-level and context-level captions through DeepSeek-R1-70B \cite{deepseekai2025deepseekr1incentivizingreasoningcapability}, which synthesizes both inputs into a unified {\em fine-grained} expression. For example, if the object-level caption states "a decorative lantern" and the context-level caption notes "the leftmost object on a mantelpiece," DeepSeek-R1-70B merges these into: "The object is a decorative lantern with intricate carvings, positioned as the leftmost item on a wooden mantelpiece flanked by framed portraits." This hierarchical approach ensures descriptions are both precise and contextually grounded.

\subsection{Model structure}
Our PixCLIP implements subtle structural modifications to the CLIP image encoder to preserve CLIP's prior knowledge. Similar to how an image is encoded by a patch embedding layer in vision transformers (ViTs)\cite{dosovitskiy2021imageworth16x16words}, we introduce a paralleled mask patch embedding layer to takes in 2D inputs with one channel. We process the input images and masks separately through respective patch embedding layers, then sum their outputs. Specifically, for the inputs images $I$ and Masks $M$, we have the feature $F$:
\begin{equation}
   \mathbf{F} =  E_n( Conv_I(I) + Conv_M(M) + P )
\end{equation}
where $\mathbf{Conv}_I(\cdot)$ and $\mathbf{Conv}_M(\cdot)$ are the image and mask patch embedding layer. Correspondingly,  $E_n$ is the vision encoder and $P$ denotes the positional encoding. The newly added mask embedding layer $\mathbf{Conv}_M(\cdot)$ is initialized to output zeros, ensuring that the model's initial behavior is unaffected prior to fine-tuning.

\begin{figure*}[bht]
  \centering
  \includegraphics[width = 0.86\textwidth]{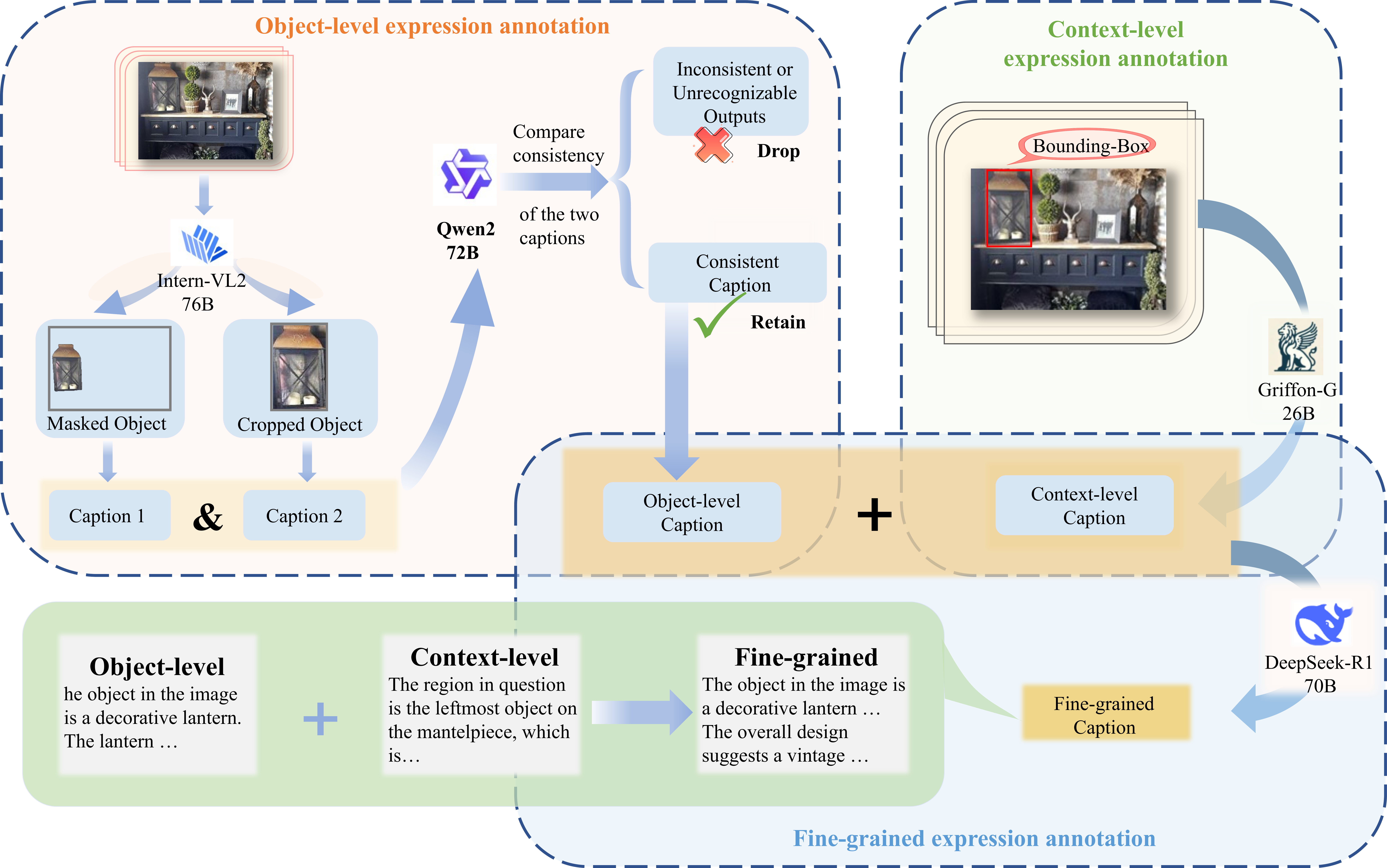}
  \caption{The pipeline of our data generation method. Using multiple MLLMs, we generate object captions and in-context captions. They are then checked and merged into fine-grained captions for use in further training stage.}
  \label{fig:data_pipeline}
\end{figure*}

\subsection{Multi-granularity Training method}

\textbf{Overview.} PixCLIP employs three main alignment strategies: global image-text alignment, local mask-text alignment, and multi-scale feature enhancement. As depicted in Fig.\ref{pipeline}, PixCLIP adopts the architecture likes CLIP, which consists of the vision encoder $V$ and the language encoder $L$, but employs more complex inputs and objectives. We initialized the ViT weights from LLM2CLIP \cite{huang2024llm2clip} and use LLAMA3-8B \cite{grattafiori2024llama3herdmodels} as our text encoder to receive long text input and leverage the open-world knowledge of LLM. Inheriting weights of LLAMA3-8B and the adaptor (a simple projector)  from LLM2Vec \cite{behnamghader2024llm2vec}, we first make self-supervised fine-tuning \cite{behnamghader2024llm2vec} on our text data following its setting and then freezing it as the text tower to reduce training overhead.
For training, each input batch consists of $\mathcal{B}$ samples. The input includes an original image $I_i$, a mask $M_i$ , and a long text $T_i$ describing the masked region. Additionally, from the original image $I_i$ and mask $M_i$, we generate a \textit{cropped image} $I'_i$ that tightly contains the masked region $M_i$, and a mask $M'_i$ outlining the object within the cropped image $I'_i$, as illustrated in Figure~3. Thus, the full input for sample $i$ is the tuple $(I_i, M_i, T_i, I'_i, M'_i)$.The long texts $T_i$ are sourced from our LongGRIT dataset.

\noindent\textbf{Mask-text contrastive learning.} 
We basically following the previous CLIP-based settings. For the inputs image and masks $\{I_i,M_i\}$, PixCLIP first encodes them to produce a visual embedding $e_v$ which is focused on the local region referred to by the mask. And LLM encodes the long text to text embedding $e_t$, the cosine similarity is calculated as:
\begin{equation}
    S(\mathbf{e_v}, \mathbf{e_t}) = \frac{\mathbf{e_v} \cdot \mathbf{e_t}}{\|\mathbf{e_v}\| \|\mathbf{e_t}\|}
\end{equation}
Model is forced to learn visual and text emebedding by maximizing the cosine similarity to the corresponding text and image embeddings, while minimizing the cosine similarity to other non-corresponding ones in the batch:
\begin{align}
    {L}_\text{CL}(\mathbf{e}_v, \mathbf{e}_t) = 
    -\frac{1}{2\mathcal{B}} &
    \sum_{i=1}^{\mathcal{B}} 
    \bigg( 
        \log \frac{\exp(S(\mathbf{e}_v^i, \mathbf{e}_t^i) / \tau)}{\sum_{j=1}^{\mathcal{B}} \exp(S(\mathbf{e}_v^i, \mathbf{e}_t^j) / \tau)} 
        \nonumber \\
        &+ \log \frac{\exp(S(\mathbf{e}_t^i, \mathbf{e}_v^i) / \tau)}{\sum_{j=1}^{\mathcal{B}} \exp(S(\mathbf{e}_t^i, \mathbf{e}_v^j) / \tau)} 
    \bigg)
\end{align}

Following previous work \cite{sun2024alpha}, we  set a 10\% ratio to occasionally set the mask to all 1 (representing the full image) to maintain the global embedding ability for the entire image. When it occurs, the text is changed to the whole image caption for global alignment.

\noindent\textbf{Fine-graind cropping alignment.} Some prior methods extract local features from global image representations, often leading to loss of fine-grained details—particularly for small objects in complex scenes. Therefore, to enhance features that are more focused on the region indicated by the mask, on another branch, we cropped the region corresponding to the mask. The cropped region tightly encloses the mask, making the content within the mask the primary focus of the image. After being enlarged, the embedding $v_c, t_c$obtained from the cropped image $I_c$ and $M_c$ via the visual encoder is aligned with the original visual embedding, to force the model to focus more on the visual details themselves:
\begin{equation}
    L_{FC} = L_{CL(\mathbf{v}_c, \mathbf{t}_c)}
\end{equation}

Since we want the main embedding to also retain contextual interaction and boundary information, a full alignment with the embedding from the cropped detailed mask is not appropriate. We can only maintain the positive sample matching relationship. This method can also prevent poor model training, keeping the model from focusing too much on the whole image when a mask is provided as input.

\noindent\textbf{Local-Global representation enhancement.} A model's ability to understand the whole image and its local
regions is complementary, meaning their optimization directions tend to converge. A correct global understanding of an image, in fact, stems precisely from the cumulative accurate understanding of all its local regions. This, in turn, dictates whether the model's contextual comprehension of any arbitrary local area is accurate. Therefore, we utilize a branch to explicitly connect these two aspects, combining them into a joint learning task. This branch enables the model to maintain its full-image capabilities and further explore the deep relationship between comprehensive image understanding and arbitrary local comprehension, ensuring both are optimized synergistically. 

Specifically, on another branch, we input the original image $I$ with an all-1 mask $M_1$. From the global dense representation, we  extract local representations once more. Specifically, regional visual representations $f'_v$ are extracted by pooling visual dense features $\{ f_v^i \}_{i \leq \mathcal{P}}$ according to the region occupied by the mask $M$, using Average Pooling:

\begin{equation}
f'_v = 
\mathit{proj}\left(\text{AvgPool}\left(\{ f_v^i \mid p_i \in \mathcal{M} \cap \mathcal{P} \text{,} \dfrac{|\mathcal{M} \cap \mathcal{P}|}{|\mathcal{P}|} > \tau \}\right) \right)
\end{equation}
\begin{equation}
    L_{LG} = L_{CL(f'_v, f_v)}
\end{equation}
Then the self-supervision process can continuously update and improve itself.

Finally the total loss is calculated as:
\begin{equation}
    L = L_{CL} + \alpha L_{FC} +\beta L_{LG}
\end{equation}
Noted that throughout the entire training process, short texts are also included and trained together with long texts, in order to enhance the model's robustness.



\begin{table*}[bth]
\centering
\begin{tabular}{lccc}
\toprule
\textbf{Methods} & \textbf{Input type} & \textbf{top1} & \textbf{top5} \\
\midrule
CLIP\cite{radford2021learningtransferablevisualmodels} & Crop Image & 66.48\% & 88.90\% \\ 
MaskedAdaptedCLIP\cite{liang2023open} & Crop Image & 57.86\% & 79.12\% \\ 
Red Circle\cite{shtedritski2023does} & Visual Prompt & 65.37\% & 88.68\% \\ 
MaskCLIP\cite{dong2023maskclip} & Visual Prompt & 67.86\% & 89.40\% \\ 
Alpha-CLIP \cite{sun2024alpha} & Visual Prompt & 68.89\% & 90.51\% \\
\rowcolor{blue!15}\textbf{PixCLIP} & \textbf{Visual Prompt} & \textbf{69.57\%} & \textbf{91.17\%} \\
\bottomrule
\end{tabular}
\caption{Accuracy comparison of Zero-shot classification results on ImageNet-S. All methods use ViT-B/16.}
\label{tab:imagenets}
\end{table*}

\begin{table*}[hbt]
    \centering
    \begin{tabular}{l|cc|cc|c}
        \toprule
        \multirow{2}{*}{\textbf{Methods}} & \multicolumn{2}{c|}{\textbf{RefCOCO}} & \multicolumn{2}{c|}{\textbf{RefCOCO+}} & \multirow{2}{*}{\textbf{RefCOCOg}} \\
         & Val & Test & Val & Test &  \\
        \midrule
        CPT\cite{yao2022cptcolorfulprompttuning} & 32.2 & 36.1 & 31.9 & 35.2 & 36.7 \\
        ReCLIP\cite{subramanian2022reclip} & 45.8 & 46.1 & 47.9 & 50.1 & 59.3 \\
        Red Circle\cite{shtedritski2023does} & 49.8 & 58.6 & 55.3 & 63.9 & 59.4 \\
        Alpha-CLIP \cite{sun2024alpha} & 55.7 & 61.1 & 55.6 & 62.7 & 61.2 \\
        \rowcolor{blue!15}\textbf{PixCLIP} & \textbf{59.93} & \textbf{69.9} & \textbf{59.1} & \textbf{68.5} & \textbf{61.8} \\
        \bottomrule
    \end{tabular}
    \caption{Performance on RefCOCO/+/g Datasets. All models use ViT-B/16.}
    \label{tab:performance_modified}
\end{table*}

\section{Experiment}

\subsection{Experimental Setting} Our experiments evaluated PixCLIP's metrics across multiple dimensions. For evaluation, we used ViT-B/16 models. Our models were trained on the LongGRIT dataset we made. The batch size was 1024 for ViT-B/16, the $\alpha$ and $\beta$ is set to 0.25. The entire training was completed on 8 H20 100G, totaling 8 epochs. 

We compared our approach to previous state-of-the-art (SOTA) models on both pixel-level regional tasks and traditional image-level tasks. For regional tasks, we selected previous works that focused on adapting CLIP for specific areas, including MaskCLIP~\cite{dong2023maskclip}, MaskAdaptedCLIP~\cite{liang2023open}, Red Circle~\cite{shtedritski2023does}, Alpha-CLIP~\cite{sun2024alpha}, and so on. We were unable to compare our method with CLOC~\cite{chen2025contrastivelocalizedlanguageimagepretraining} as it has not yet open-sourced its model. In traditional image-text retrieval, our method showed strong performance on both short- and long-text datasets, outperforming models specifically designed for long-text retrieval, such as LongCLIP~\cite{zhang2024long} and LLM2CLIP~\cite{huang2024llm2clip}.

\begin{table}[htb]
\centering

\begin{tabular}{lcc}
\hline
\multirow{2}{*}{\textbf{Methods}} & \multicolumn{2}{c}{\textbf{Zero-shot Classification}} \\
\cline{2-3} 
 & Top1 & Top5 \\ \hline
CLIP\cite{radford2021learningtransferablevisualmodels} & 64.21\% & 86.69\% \\ 
Alpha-CLIP \cite{sun2024alpha}& 71.08\% & 88.90\% \\ 
LLM2CLIP \cite{huang2024llm2clip}& 70.16\% & 87.40\% \\
\rowcolor{blue!15} \textbf{PixCLIP} & \textbf{76.68\%} & \textbf{90.56\%} \\ 
        \bottomrule
\end{tabular}
\caption{Accuracy comparison on \textbf{Instance-COCO}. All methods use ViT-B/16.}
\label{tab:coco}
\end{table}

\subsection{PixCLIP in Region Recognition}

\textbf{Zero-shot region classification.} We firstly evaluate our model in ImageNet-S \cite{Gao_2023} dataset for zero-shot classification analysis, which comprises 919 classes with semantic segmentation annotations selected from ImageNet-1k. The result is shown as Table.\ref{tab:imagenets}:

For scenarios that need to crop or mask objects in images \cite{subramanian2022reclip,chen2023ovarnet} . Previous work \cite{sun2024alpha} has conducted classification tests for in such scenarios using the validation set of the Instance-COCO \cite{lin2015microsoftcococommonobjects} dataset, which consists of 80 classes. We followed the setting to crop objects using ground-truth bounding boxes and enlarged them by 1.5 times (referred to as coco crop). The experimental results are as Table.\ref{tab:coco}. Importantly, CLIP and LLM2CLIP take the cropped images directly as input.

\noindent\textbf{Zero shot REC(referring expression comprehension).} For the REC task, we conducted experiments on RefCOCO\cite{lin2015microsoftcococommonobjects}, RefCOCO+\cite{lin2015microsoftcococommonobjects}, and RefCOCOg\cite{lin2015microsoftcococommonobjects}. Following the setup of Alpha CLIP, we used object proposals predicted by a pretrained detector \cite{yu2018mattnet} and employed SAM to obtain masks for each proposal. On this task, we significantly outperformed the previous SOTA.

\subsection{PixCLIP in Multi-modal Retrieval.} 

As a CLIP-like multimodal foundation model, we meticulously evaluated PixCLIP's performance on cross-modal retrieval tasks. This evaluation included multiple traditional image-text retrieval benchmarks to assess our model's retention of holistic image understanding, alongside our newly extended pixel-level benchmark: Mask and Text Retrieval. Experimental results clearly show PixCLIP delivers state-of-the-art (SOTA) performance, whether dealing with full images or localized regions, and whether processing concise or extended text descriptions.

\begin{figure}[h]
  \centering
  \includegraphics[scale=0.12]{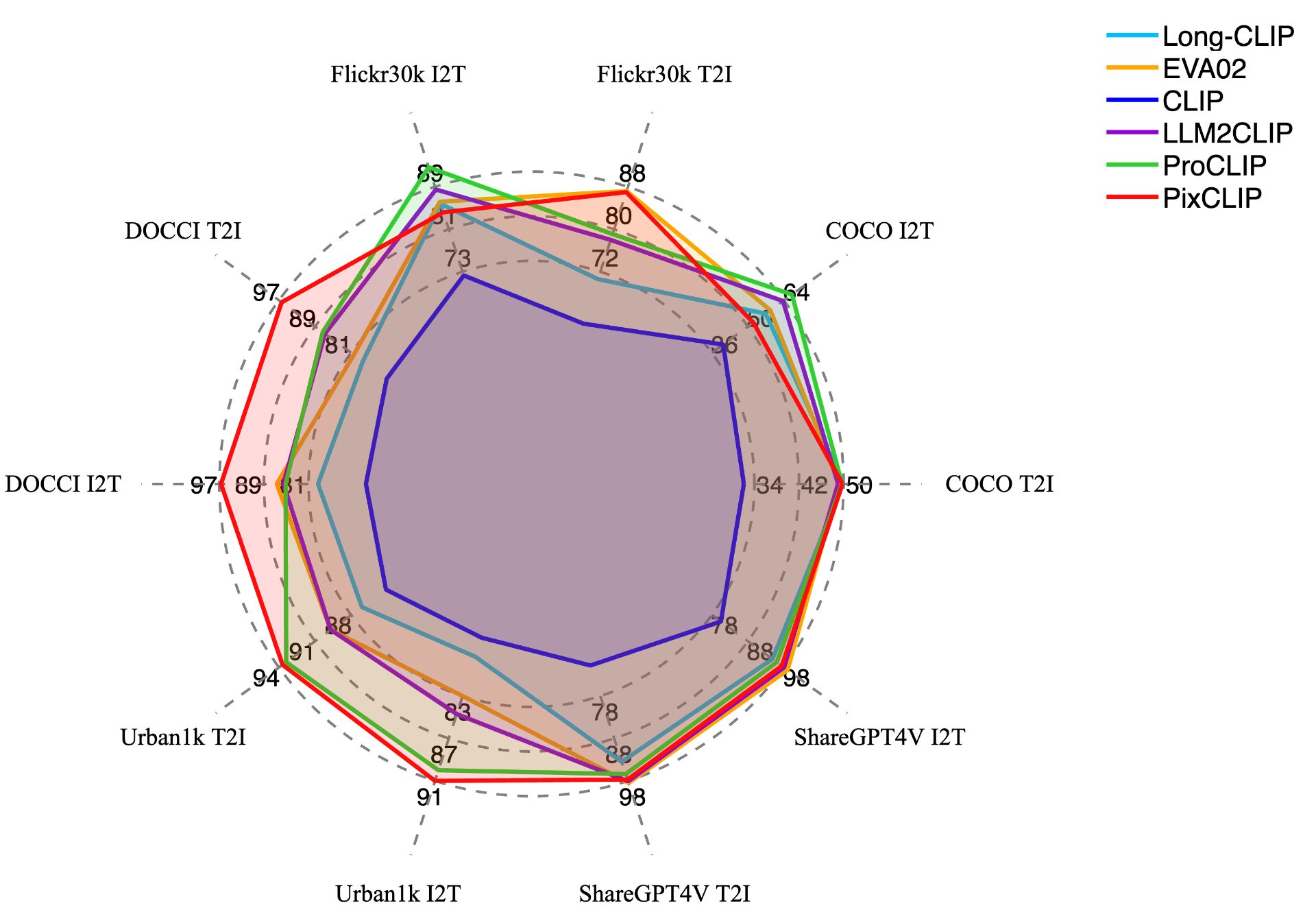}
  \caption{Retrieval comparison with previous models in traditional Image-Text benchmarks.}
  \label{fig:image-text retrieval}
\vspace{-12pt}
\end{figure}

\noindent\textbf{Traditional Retrieval with Full-Image and Text.}  For short-text retrieval, we used the MSCOCO \cite{lin2015microsoftcococommonobjects} 5K test set and the Flickr \cite{young2014image} 1K test set. For long-text retrieval, we employed datasets from LongCLIP, including a 1K subset of ShareGPT4V \cite{chen2024sharegpt4videoimprovingvideounderstanding}, the Urban1K \cite{zhang2024long} dataset, and the DOCCI \cite{onoe2024docci} dataset. The ShareGPT4V-1M dataset consists of captions generated using GPT-4V and ShareCaptioner, covering images from LAION, CC, SBU \cite{ordonez2011im2text}, and MS COCO. Urban1K includes captions for 1,000 urban scene images, each richly annotated with detailed descriptions. DOCCI contains 1.5K high-resolution images with human-annotated captions and was used for retrieval evaluation. Results are shown as Table.\ref{tab:retrieval_performance} and Fig.\ref{fig:image-text retrieval}.

\begin{table*}[h] 
\centering

\begin{tabular}{l *{12}{c}}
\toprule
\multirow{2}{*}{\textbf{Methods}} & \multicolumn{2}{c}{\textbf{Flickr30k}} & \multicolumn{2}{c}{\textbf{COCO}} & \multicolumn{2}{c}{\textbf{ShareGPT4V}} & \multicolumn{2}{c}{\textbf{Urban-1k}} & \multicolumn{2}{c}{\textbf{DOCCI}} & \multicolumn{2}{c}{\textbf{Average}} \\
        \cmidrule(lr){2-3} \cmidrule(lr){4-5} \cmidrule(lr){6-7} \cmidrule(lr){8-9} \cmidrule(lr){10-11} \cmidrule(lr){12-13}
& I2T & T2I & I2T & T2I & I2T & T2I & I2T & T2I & I2T & T2I & I2T & T2I \\
\midrule
ALIGN & 80.6 & 62.2 & 52.0 & 43.2 & 75.9 & 80.6 & 62.2 & 59.1 & 59.7 & 62.1 & 66.1 & 61.4 \\
BLIP & 80.6 & 74.1 & 61.7 & 48.5 & 65.8 & 74.3 & 45.5 & 48.5 & 50.5 & 53.5 & 60.8 & 59.8 \\ 
Jina-CLIP & 80.6 & 67.4 & 55.6 & 41.1 & - & - & 87.7 & 88.0 & 78.7 & 80.0 & 75.7 & 69.1 \\
Long-CLIP & 85.8 & 70.6 & 56.9 & 40.9 & 94.8 & 93.5 & 79.1 & 79.1 & 63.1 & 71.4 & 75.9 & 71.1 \\
CLIP & 82.3 & 62.2 & 52.4 & 33.1 & 84.5 & 79.8 & 67.5 & 53.1 & 60.7 & 57.1 & 69.5 & 57.1 \\
EVA02 & 86.2 & 71.5 & 58.7 & 42.1 & 90.5 & 85.5 & 67.0 & 60.8 & 67.7 & 68.0 & 74.0 & 65.6 \\
\rowcolor{orange!15}LLM2CLIP & \textbf{88.5} & 78.0 & \textbf{63.6} & 49.8 & \textbf{98.0} & \textbf{98.1} & 84.7 & 89.7 & 85.5 & 86.8 & 84.1 & 80.5 \\

\rowcolor{blue!15}\textbf{PixCLIP} & 84.2 & \textbf{87.0} & 52.7 & \textbf{50.2} & 97.4 & 97.7 & \textbf{91.0} & \textbf{93.7} & \textbf{96.7} & \textbf{96.4} & \textbf{84.4} & \textbf{85.0} \\
\bottomrule
\end{tabular}
\caption{Retrieval performance comparison on existing datasets.}
\label{tab:retrieval_performance}
\end{table*}



\begin{table*}[htb]
    \centering
    \begin{tabular}{lcccccc}
        \toprule
        \multirow{2}{*}{Methods} & \multicolumn{3}{c}{M2T} & \multicolumn{3}{c}{T2M} \\
        \cmidrule(lr){2-4} \cmidrule(lr){5-7}
        & Recall@1 & Recall@5 & Recall@10 & Recall@1 & Recall@5 & Recall@10 \\
        \midrule
        Alpha CLIP\cite{sun2024alpha} & 28.3\% & 50.4\% & 59.9\% & 27.5\% & 46.9\% & 55.7\% \\
        
        \rowcolor{blue!15}\textbf{PixCLIP} & \textbf{47.3\%} & \textbf{66.4\%} & \textbf{73.4\%} & \textbf{47.9\%} & \textbf{66.8\%} & \textbf{74.1\%} \\
        \bottomrule
    \end{tabular}
    \caption{Accuracy comparison on Zero-shot mask-to-text (M2T) and text-to-mask (T2M) results on Ref-SAV.}   
        \label{tab:refsav}
\end{table*}

    

\begin{table*}[h!]
    \centering
 
    \begin{tabular}{lccccc}
        \toprule
        \multirow{2}{*}{\textbf{Models}} & \multicolumn{3}{c}{\large\textbf{Region Tasks}} & \multicolumn{2}{c}{\large\textbf{Full-Image Tasks}} \\
        \cmidrule(lr){2-4} \cmidrule(lr){5-6}
        & RefCOCO Val & Ref-SAV M2T & Ref-SAV T2M & Urban1k I2T & Urban1k T2I\\
        \midrule
        LLM2CLIP\cite{huang2024llm2clip} & - & - & - & 84.7\% & 89.7\% \\
        $L_{CL}$ & 51.144\% & 47.0\% & 46.9\% & 87.1\% & 92.4\% \\
        $L_{CL}+L_{LG}$ & 59.041\% & \textbf{47.4\%} & 47.3\% & 88.9\% & 93.2\% \\
        
        \rowcolor{orange!15}\textbf{$L_{CL}+L_{LG}+L_{FC}$} & \textbf{59.926\%} & 47.3\% & \textbf{47.9\%} & \textbf{91.0\%} & \textbf{93.7\%} \\
        \bottomrule
    \end{tabular}
      \caption{The ablation results on both pixel-level and image-level tasks.}
    \label{tab:ab_combined}  
\end{table*}

\noindent\textbf{Pixel-level Retrieval with Mask and Text.}  Ref-SAV \cite{yuan2025sa2va} is a dataset originally intended for video referring segmentation. Ref-SAV includes 70,000 masklets and corresponding long captions, with an average caption length of over 100 words, providing fine-grained details. Although initially designed for video referring segmentation, we found it particularly well-suited for evaluating our model's performance. We extracted 1,000 mask-text samples for retrieval and compared our work with the previous SOTA work Alpha-CLIP. Our work significantly outperforms Alpha-CLIP in performance. This is primarily due to our LLM's ability to better extract fine-grained text features from longer captions, a capability fundamentally lacking in Alpha-CLIP's traditional text encoder.

\vspace{2pt}

\subsection{Ablation Experiment}
The PixCLIP training objective consists of three components: $L_{CL}$, $L_{FC}, L_{LG}$. We conducted the ablation study on these components across both \textbf{regional-level tasks} (REC and mask-text retrieval on Ref-SAV) and \textbf{image-level tasks}.

Our experiments show that simply using direct alignment between mask-based visual features and text is inefficient, leading to poor performance and degradation on whole-image inputs. We hypothesize this is because, while the LLM provides rich text features, the limited amount of local data makes it challenging for the model to learn robust representations that work for both local and global visual inputs. The $L_{FC}, L_{LG}$ successfully resolves this issue and significantly boosts performance.

This result precisely confirms the hypothesis we outlined in our method: a strong foundation model's capability for regional and broad enhance each other.

\section{Conclusion}
In this paper, we propose PixCLIP, a model capable of accepting arbitrary local input on the image side indicated by a mask and processing text input of arbitrary length on the text side, to achieve deeper, multi-level image-text alignment. For significance, we are no longer limited to traditional monotonous image-text pairs, achieving perfect adaptation to arbitrary image pixels and long text sequences, thus being a more unified CLIP. For training, we propose a multi-grained alignment framework with two additional branches: Fine-grained Cropping Alignment and Local-Global Representation Enhancement, for performance enhancement. For data, we engineered an automatic pipeline to construct a dataset of 1.5M masks with corresponding fine-grained captions, serving as a benchmark for mask-text retrieval tasks. We believe this research will facilitate further exploration in the multimodal domain.

\bibliography{reference}

\newpage
\clearpage
\appendix
\section{More Evaluation Results}
Due to space limitations in the main manuscript, we further present the experimental results for ViT-L/14 here. We tested and report PixCLIP's performance on Instance-COCO and ImageNet-S.

\begin{figure}[h!]
  \centering
  \includegraphics[scale=0.28]{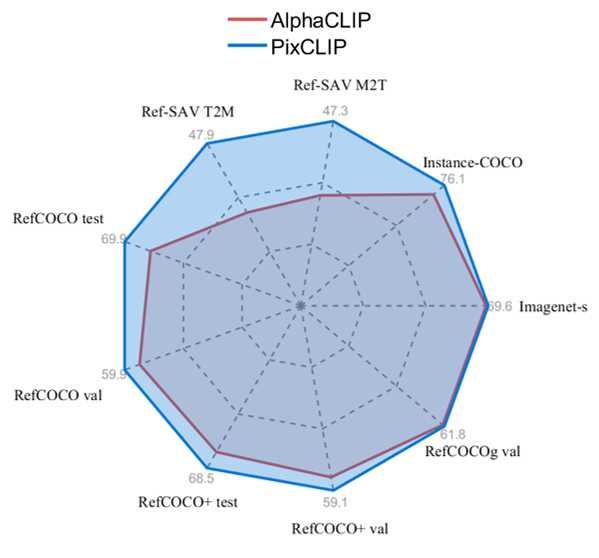}
  \caption{Performance comparison between PixCLIP and prior state-of-the-art works.}
  \label{fig:sup_circle}
\end{figure}

The results for zero-shot region classification is shown as Table.\ref{tab:imagenets} and Table.\ref{tab:coco}. The experiments effectively demonstrate that when provided with a foreground object mask through the alpha channel, our PixCLIP generates visual features that are more focused on the foreground object, leading to better image-level classification compared to the original CLIP and other baseline approaches. Notably, we followed the settings of Alpha-CLIP to enable MaskCLIP to accept task-specific formats. We also visualized and demonstrated PixCLIP's leading performance over previous SOTA work AlphaCLIP on various benchmarks, as shown in Figure.\ref{fig:sup_circle}. Furthermore, we compared the indicators of our method with those of the most advanced approach on the fine-grained analysis benchmark. The results are shown in Table \ref{tab:fine-granied}.

\begin{table}[htb]
\centering
\begin{tabular}{l|c|cc}
\hline
\multirow{2}{*}{\textbf{Methods}} & \multirow{2}{*}{\textbf{Input}} & \multicolumn{2}{c}{\textbf{Imagenet-S }} \\
\cline{3-4}
 & & Top1 & Top5 \\
\hline
\rowcolor{gray!8} \multicolumn{4}{l}{\textbf{ViT-B/16}} \\
\hline
CLIP & C & 66.48\% & 88.90\% \\
MaskedAdaptedCLIP & C & 57.86\% & 79.12\% \\
Red Circle & VP & 65.37\% & 88.68\% \\
MaskCLIP & VP & 67.86\% & 89.40\% \\
Alpha-CLIP & VP & 68.89\% & 90.51\% \\
\rowcolor{blue!15} \textbf{PixCLIP} & \textbf{VP} & \textbf{69.57\%} & \textbf{91.17\%} \\
\hline
\rowcolor{gray!8} \multicolumn{4}{l}{\textbf{ViT-L/14}} \\
\hline
CLIP & C & 73.48\% & 91.60\% \\
MaskedAdaptedCLIP & C & 73.37\% & 92.09\% \\
Red Circle & VP & 77.04\% & 93.39\% \\
MaskCLIP & VP & 77.04\% & 93.39\% \\
Alpha-CLIP & VP & 77.41\% & 94.45\% \\
\rowcolor{blue!15} \textbf{PixCLIP} & \textbf{VP} & \textbf{79.42\%} & \textbf{94.77\%} \\
\hline
\end{tabular}
\caption{Accuracy comparison of zero-shot classification on \textbf{ImageNet-S}. The "C" in the table signifies models that take a mask's corresponding \textbf{Cropped Image} as direct input. "VP" indicates that, in addition to the image, the model can also directly accept a \textbf{Visual Prompt}. }
\label{tab:imagenets}
\end{table}

\begin{table}[htb]
\centering

\begin{tabular}{lcc}
\hline
\multirow{2}{*}{\textbf{Methods}} & \multicolumn{2}{c}{\textbf{Zero-shot Classification}} \\
\cline{2-3} 
 & Top1 & Top5 \\ \hline
\rowcolor{gray!8}\multicolumn{3}{l}{\textbf{ViT-B/16}} \\ 
\hline
CLIP & 64.21\% & 86.69\% \\ 
Alpha-CLIP & 71.08\% & 88.90\% \\ 
LLM2CLIP & 70.16\% & 87.40\% \\
\rowcolor{blue!15} \textbf{PixCLIP} & \textbf{76.68\%} & \textbf{90.56\%} \\ 
        \bottomrule
\rowcolor{gray!8}\multicolumn{3}{l}{\textbf{ViT-L/14}} \\ 
\hline
CLIP & 71.48\% & 90.15\% \\ 
Alpha-CLIP & 77.48\% & 94.40\% \\ \hline
LLM2CLIP & xx.xx\% & xx.xx\% \\ \hline
\rowcolor{blue!15} \textbf{PixCLIP} & \textbf{79.42\%} & \textbf{94.77\%} \\ \hline
\end{tabular}
\caption{Accuracy comparison on \textbf{Instance-COCO}.}
\label{tab:coco}
\end{table}



\begin{table}[htb]
\centering
\renewcommand{\arraystretch}{1.2}
\begin{tabular}{@{}lc@{}}
\toprule
\textbf{Parameter} & \textbf{Value} \\
\midrule
Model Structure      & EVA02 \\
Backbone             & ViT-B/16 \\
Optimizer            & AdamW \\
Learning Rate        & $1 \times 10^{-5}$ (Mask Conv) \\
                     & $1 \times 10^{-7}$ (Other Layers) \\
$\alpha$             & 0.25 \\
$\beta$              & 0.25 \\
Warm Up Length       & 800 steps \\
Train Epochs         & 8 \\
Full Image Ratio     & 0.1 \\
Adaptor Layers       & 4 \\
Batch Size           & $128 \times 8$ \\
AMP (bf16 training)  & True \\
Clip Log Scale       & 4.0652 \\
Weight Decay         & $1 \times 10^{-2}$ \\
\bottomrule
\end{tabular}
\caption{Detailed List of Parameters for Model and Training}
\label{tab:training_parameters}
\end{table}

\section{Detailed Setting} In this subsection, we list our model architecture and all parameters pertaining to the experimental setup. The detailed parameters are shown in Table \ref{tab:training_parameters}.

\section{Inference Stage}

\begin{figure*}[h!]
 \centering
 \includegraphics[scale=0.12]{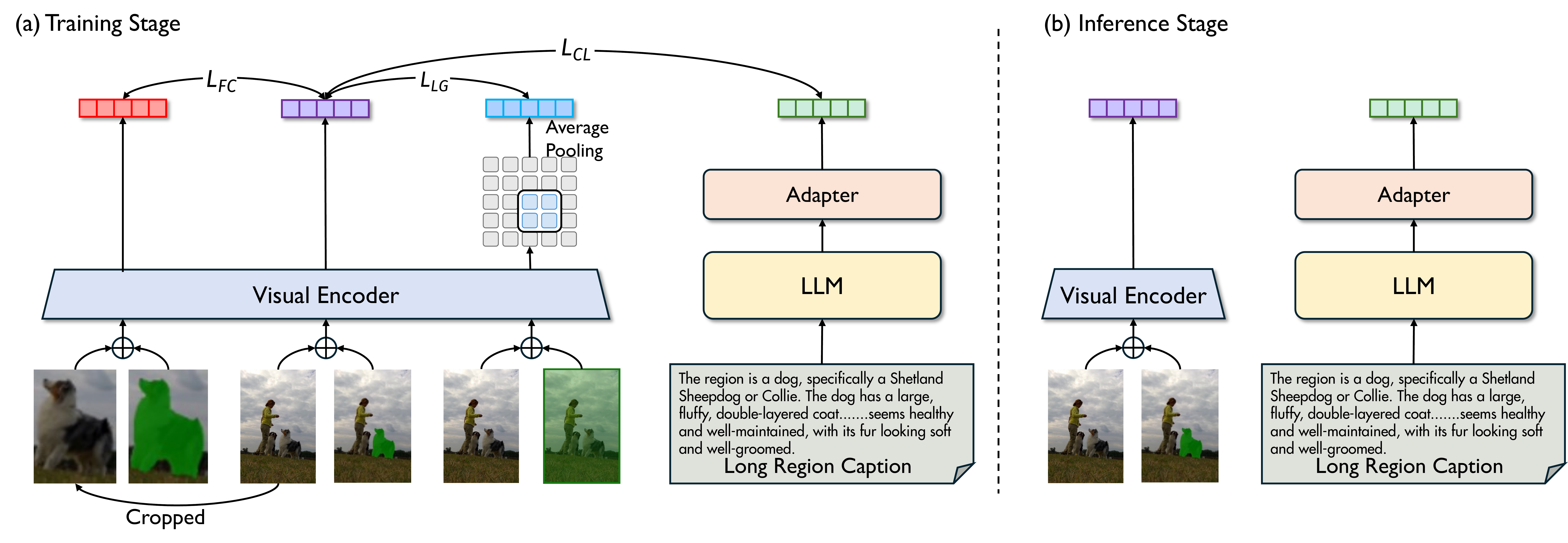}
    \caption{The distinction between PixCLIP at training time and inference time.}
\vspace{-1pt}
\label{fig:inference}
\end{figure*}

During training, we adopt a three-branch framework to train the visual encoder. In the inference stage, the visual encoder only requires the full image and the corresponding mask, without the need for cropping or extracting features from specific regions, as illustrated in Figure \ref{fig:inference}.



\definecolor{lightorange}{rgb}{0.95,0.85,0.75}  
\definecolor{lightgray}{rgb}{0.93,0.93,0.93}    
\begin{table*}[htb]
    \centering
    \begin{tabular}{l | c c c c | c c c }
    \toprule
    \multirow{2}{*}{\textbf{Model}} & \multicolumn{2}{c}{\cellcolor{lightorange}\textbf{MSCOCO}} & \multicolumn{2}{c|}{\cellcolor{lightorange}\textbf{Flickr30k}} & \multicolumn{3}{c}{\cellcolor{blue!15}\textbf{Winoground}} \\
    &\cellcolor{lightorange} I2T &\cellcolor{lightorange} T2I &\cellcolor{lightorange} I2T &\cellcolor{lightorange} T2I & \cellcolor{blue!15}T. & \cellcolor{blue!15}I. & \cellcolor{blue!15}G. \\
    \midrule
    \multicolumn{8}{c}{\textit{Model Architecture: ViT-B/16}} \\
    DreamLIP & 53.3 & 41.2 & 82.3 & 66.6 & 26.0 & 10.00 & 7.25 \\
    LiT & 30.0 & 16.5 & 54.8 & 38.5 & 24.3 & 6.5 & 4.8 \\
    ShareLock & 26.0 & 13.5 & 53.9 & 34.9 & 26.3 & 12.8 & 5.3 \\
    CLIP-B & 55.4 & 38.3 & 83.2 & 65.5 & 25.7 & 11.5 & 7.75 \\
    SAIL-B-GTE & 48.2 & 37.9 & 76.5 & 63.9 & 31.0 & 11.5 & 9.5 \\
    SAIL-B-NV2 & \textbf{57.3} & 45.3 & 84.1 & 70.1 & 35.0 & 17.25 & \textbf{13.0} \\
    PixCLIP & 52.7 & \textbf{50.2} & \textbf{84.2} & \textbf{87.0} & \textbf{35.25} & \textbf{17.75} & 12.0 \\
    \bottomrule
    \end{tabular}
    \caption{Results on complex-reasoning and fine-grained tasks.}
    \label{tab:fine-granied}
\end{table*}

\section{Broader Impact} Our PixCLIP model overcomes the limitations of traditional methods in vision-language tasks. While conventional models can only accept full image input and text inputs within a fixed length, our model can effectively and tightly align image regions of arbitrary granularity and texts of arbitrary length into the same feature space. We contribute to developing a coherent learning framework that enhances CLIP's fine-grained capabilities and transferability. Simultaneously, we pioneer a new task, namely retrieval between image regions and text, and provide the LongGRIT dataset for benchmarking future models. To ensure a positive social impact, the data we use does not involve any personal privacy issues.

\section{Limitation} PixCLIP still has two significant limitations: \textbf{(1)} Although we possess high-quality data, the size of data is insufficient, with only 1.5M samples, hindering effective scaling up. Future approaches might involve more extensive self-supervised training with larger amounts of unlabeled data or constructing more efficient data utilization pipelines. \textbf{(2)} When creating the LongGRIT dataset, we did not specifically consider hard negative samples—for instance, having MLLMs intentionally modify a core attribute in a correct description. Future work based on this direction might achieve higher data quality and more efficient data utilization by generating such samples.

\section{More Visualization Results} Follow the previous works, we visualize PixCLIP's attention maps to check whether our model pays more attention to user-defined highlighted areas at feature grid space. For a fair comparison,  We check the attention map of [CLS] token in the last transformer block in the vision encoder.  

\begin{figure*}[h!]
  \centering
  \includegraphics[scale=0.25]{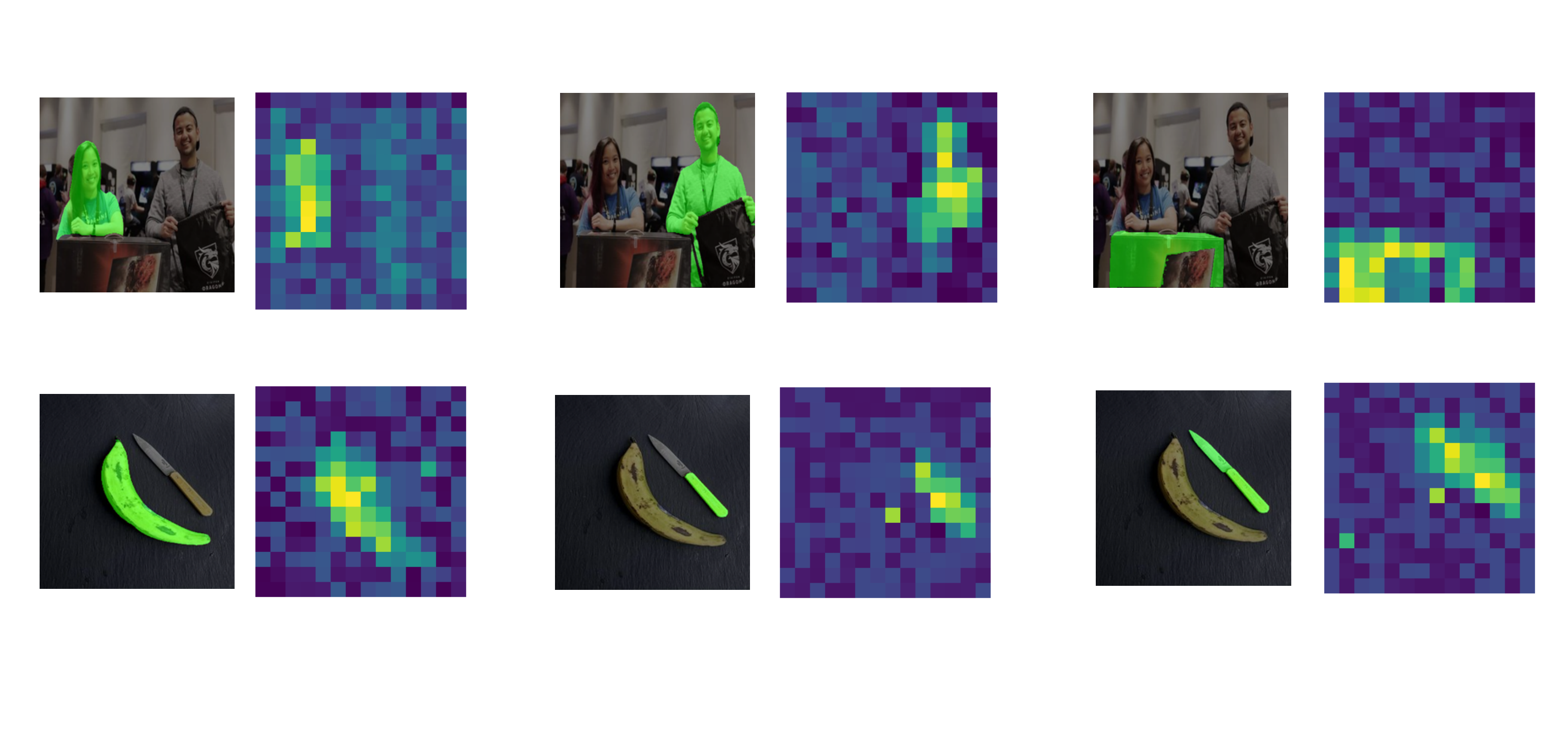}
    \caption{Visualized examples of attention map}
\vspace{-1pt}
\label{fig:attention map}
\end{figure*}

Results are shown in Fig.\ref{fig:attention map}.  This visualization verifies that our model pays more attention to the area to focus on and effectively learns to understand fine-grained semantics.

\begin{figure*}[h!]
  \centering
  \includegraphics[scale=0.21]{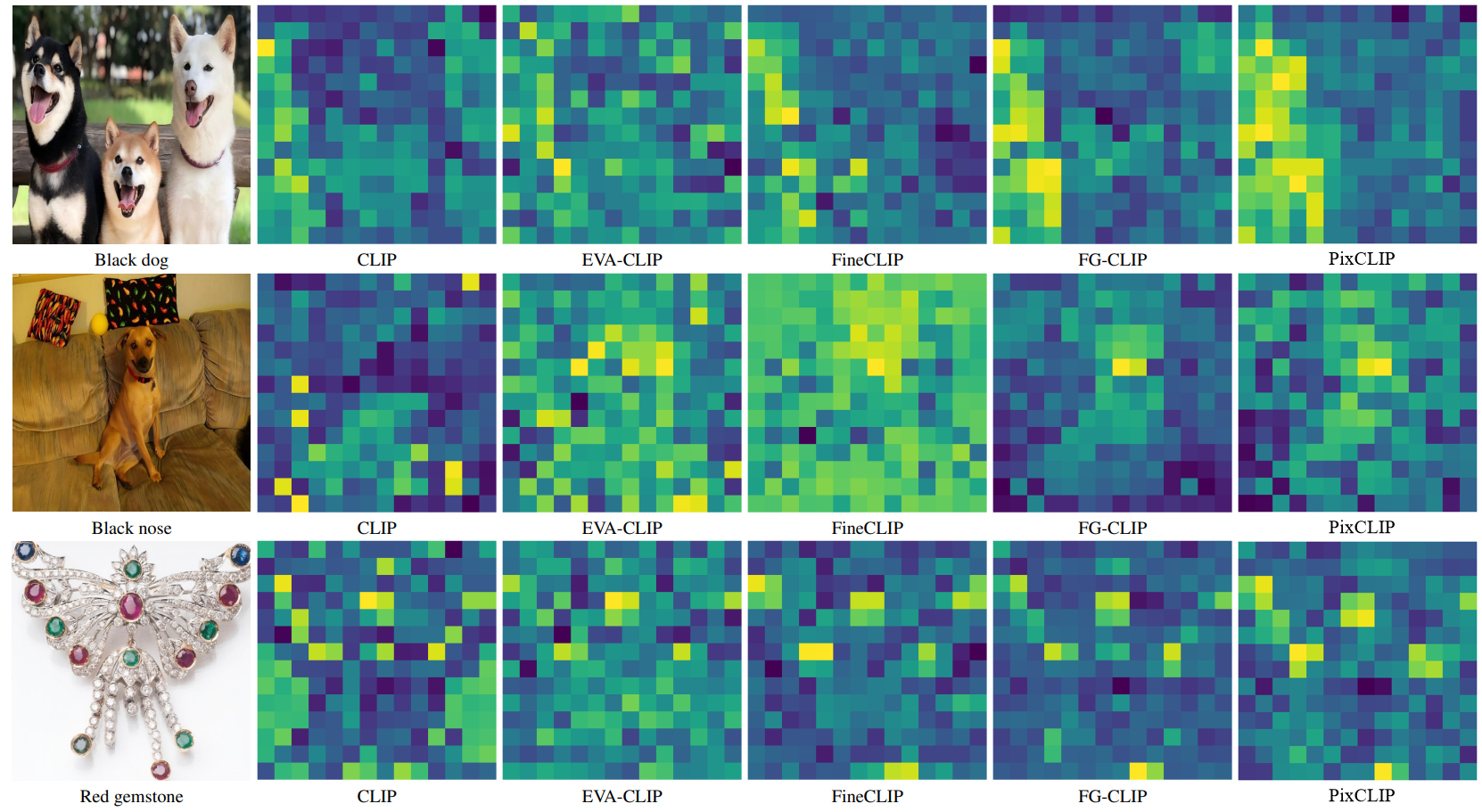}
    \caption{Visualized examples of attention map}
\vspace{-12pt}
\label{fig:full attention map}
\end{figure*}

Additionally, we visualized our model's fine-grained responsiveness to text when given a whole image as input. We computed an attention map by calculating the similarity between the feature map from our model's last layer and the text features. Fig.\ref{fig:full attention map} presents our visualization results, which demonstrate that our mask-guided input enhances local features when processing the entire image. We maintain a comparable performance level compared to previous fine-grained methods."

\section{More Examples For Data Generation} We provide some visualized examples of our LongGRIT, shown in Fig.3. Additionally, we showcase the features of our LongGRIT dataset through samples that demonstrate multi-granularity, occlusion reappearance, and both short and long text expressions. These features make LongGRIT also be a challenging and comprehensive \textbf{Mask-Text Retrieval} benchmark.

\begin{figure*}
  \centering
  \includegraphics[scale=0.40]{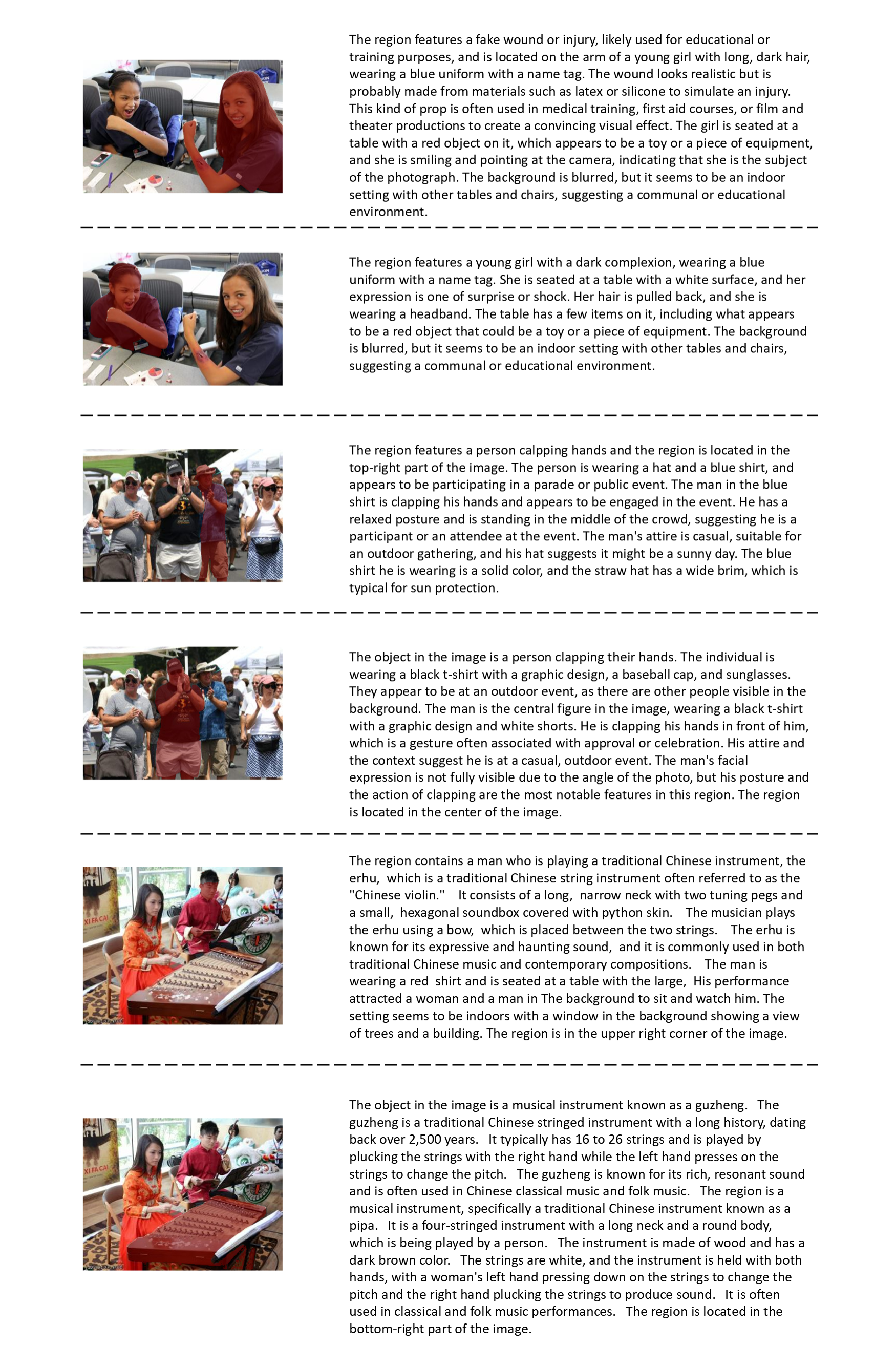}
    \caption{Visualized examples of our LongGRIT.}
\vspace{-12pt}
\end{figure*}

\begin{figure*}
  \centering
  \includegraphics[scale=0.40]{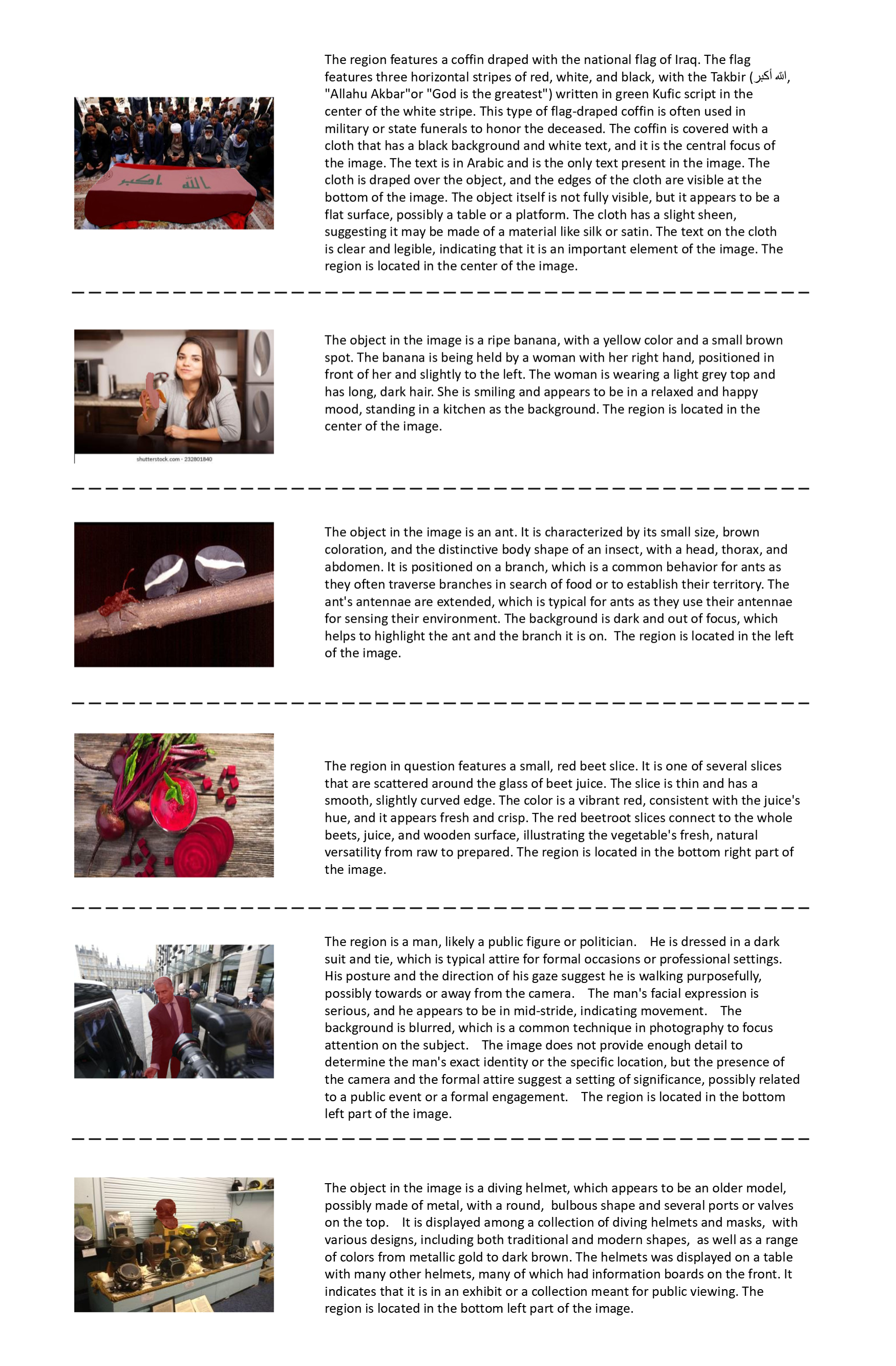}
    \caption{More visualized examples of our LongGRIT.}
\vspace{-12pt}
\end{figure*}

\begin{figure*}[h!]
  \centering
  \includegraphics[scale=0.21]{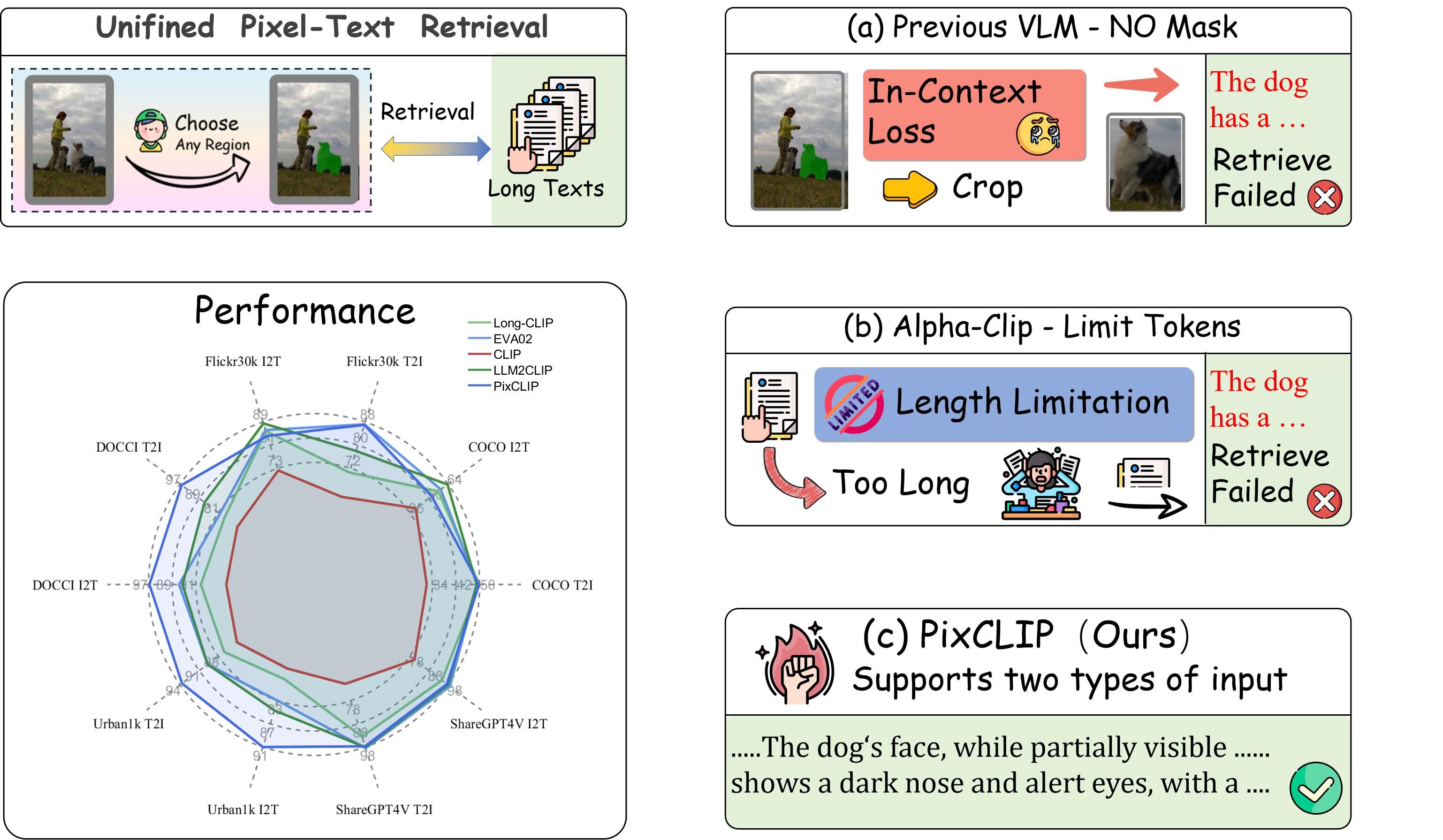}
  \caption{PixCLIP: Achieving Fine-grained Visual Language Understanding via Any-granularity Pixel-Text Alignment Learning.}
  \label{fig:first}

\end{figure*}

\vspace{-12pt}

\clearpage

\end{document}